\documentclass{article} 
\usepackage{geometry}
\usepackage[english]{babel} 
\usepackage{amssymb}
\usepackage{amsmath}
\usepackage{multirow}
\usepackage{txfonts}
\usepackage{multicol}
\usepackage{subfigure}
\usepackage{booktabs}
\usepackage{mathdots}
\usepackage{graphicx} 
\usepackage{fancyhdr}
\usepackage{hyperref}
\usepackage{float}
\usepackage{microtype}
\usepackage{caption}
\usepackage[numbers]{natbib}
\usepackage[ruled,vlined]{algorithm2e}

\makeatletter
\def\@xfootnote[#1]{%
  \protected@xdef\@thefnmark{#1}%
  \@footnotemark\@footnotetext}
\makeatother

\usepackage{sectsty}
\subsectionfont{\normalfont\itshape}
\subsubsectionfont{\normalfont\itshape}

\begin{document}


\begin{tabular}{p{1.1in}p{4.5in}p{1.2in}}  
\hspace{-1cm}
\noindent
\begin{tabular}{c}  \includegraphics[width=2.9cm]{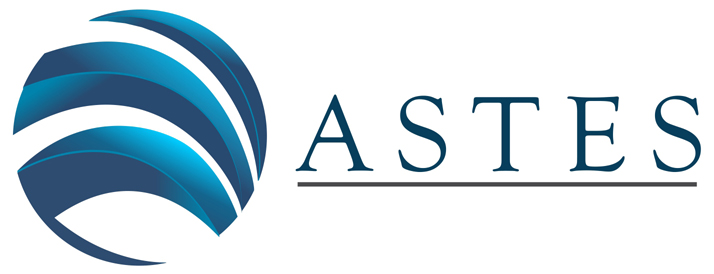}\end{tabular} 	& \vspace{-0.5cm} \centering \textit{Advances in Science, Technology and Engineering Systems Journal \newline Vol. 5, No. 5, XX-YY (2020)} \\   \href{http://www.astesj.com}{www.astesj.com}  
	& \vspace{-0.6cm}  \rule{1.2in}{1.5pt} \vspace{-0.2cm} \newline \centering \textbf{ ASTES Journal \newline ISSN: 2415-6698} \newline \rule{1.2in}{1.7pt} 
\end{tabular}

\vspace{1.8cm}

\noindent \textbf{\LARGE{\setlength\itemsep{0pt}Advanced Multiple Linear Regression Based Dark Channel Prior Applied on Dehazing Image and Generating Synthetic Haze}}

\vspace{0.2cm}

\noindent Binghan Li\footnote[*]{Binghan Li, Department of Electrical \& Computer Engineering, Texas A\&M University, College Station, TX, 77840, USA, Email: libinghan1994@outlook.com, }${}^{,1}$, Yindong Hua${}^{2}$, Mi Lu${}^{3}$   

\vspace{0.2cm}
\noindent\textit{${}^{1}$Department of Electrical \& Computer Engineering, Texas A\&M University, College Station, TX, 77840, USA}

\vspace{0.2cm}
\noindent\textit{${}^{2}$Department of Electrical \& Computer Engineering, Stony Brook University, Stony Brook, NY, 11794, USA}

\vspace{0.2cm}
\noindent\textit{${}^{3}$Department of Electrical \& Computer Engineering, Texas A\&M University, College Station, TX, 77840, USA}

\vspace{0.3cm}

\noindent\begin{tabular}{p{1.7in} p{0.1in} p{5.2in} }
A R T I C L E \hspace{0.1cm} I N F O &  & A B S T R A C T \\ 
 \cline{1-1}  \cline{3-3} \setlength\itemsep{0pt} \vspace{-0.1cm}
\textit{Article history:
	\newline Received: 25 December, 2020
	\newline Accepted: 02 March, 2021
	\newline Online: 11 March, 2021
	\newline \rule{1.78in}{0.5pt} 
	Keywords: 
	\newline Haze Removal
    \newline Dark Channel Prior
    \newline Multiple Linear Regression
    \newline Convolutional Neural Network
	\newline Object Detection
	\newline Synthetic Haze}
 \newline \newline  & & \vspace{-0.1cm} 
 \textit{Haze removal is an extremely challenging task, and object detection in the hazy environment has recently gained much attention due to the popularity of autonomous driving and traffic surveillance. In this work, the authors propose a multiple linear regression haze removal model based on a widely adopted dehazing algorithm named Dark Channel Prior. Training this model with a synthetic hazy dataset, the proposed model can reduce the unanticipated deviations generated from the rough estimations of transmission map and atmospheric light in Dark Channel Prior. To increase object detection accuracy in the hazy environment, the authors further present an algorithm to build a synthetic hazy COCO training dataset by generating the artificial haze to the MS COCO training dataset. The experimental results demonstrate that the proposed model obtains higher image quality and shares more similarity with ground truth images than most conventional pixel-based dehazing algorithms and neural network based haze-removal models. The authors also evaluate the mean average precision of Mask R-CNN when training the network with synthetic hazy COCO training dataset and preprocessing test hazy dataset by removing the haze with the proposed dehazing model. It turns out that both approaches can increase the object detection accuracy significantly and outperform most existing object detection models over hazy images. }\\
 \cline{1-1}  \cline{3-3}
\end{tabular}

\vspace{0.5cm}

\begin{multicols}{2}

\section{ Introduction}
This paper is an extension of work initially presented in conference name \cite{li2018multiple}. Computer vision has recently played a major role in broad applications on urban traffic, such as autonomous and assisted driving, traffic surveillance, and security maintenance. However, the existence of haze, mist, dust, and fumes can severely degrade the visibility of images captured outside. Haze generates reduced contrasts, fainted surfaces, and color distortion to outdoor scenes, which will inevitably complicate many advanced computer vision tasks, including object classification and segmentation. Since the depth information of haze is non-linear and dependent over a global scene, haze removal becomes a challenging task. Most computer vision algorithms are designed based upon haze-free input images. They benefit a lot from haze removal, making it a highly desired task in computational photography and computer vision applications. 

Many algorithms have been proposed to restore clear images from hazy images. Polarization-based methods \cite{schechner2001instant} \cite{shwartz2006blind} \cite{lu2004arithmetic} analyze the polarization effects of atmospheric scattering and remove haze through as few as two images with different degrees of polarization. Depth-based approaches \cite{kopf2008deep} \cite{narasimhan2003interactive} evaluate depth information upon some assumptions or priors, then estimate transmission map $t(x)$ and atmospheric light $A$ from it. Some recent CNN-based haze-removal models \cite{li2017all} \cite{cai2016dehazenet} \cite{ren2016single} are built upon various powerful CNNs to self-learn transmission map $t(x)$ directly from large-scale image datasets. Among effective conventional dehazing algorithms, Dark Channel Prior (DCP) \cite{he2010single} is generally accepted due to its novel prior and outstanding performance. In most non-sky patches of the haze-free image, at least one color channel contains dark pixels with extremely low intensity, which is primarily generated by the air light. However, the estimation on the medium transmission $t(x)$ and atmospheric light $A$ is not precise, especially when the scene object is inherently similar to the air light over a large local region and no shadow is cast on it. And the restored image looks unnaturally dark when there is a sky region with sunlight.

In this paper, the authors propose a novel Multiple Linear Regression Dark Channel Prior based model (MLDCP). Trained with the training dataset in REalistic Single Image DEhazing (RESIDE) \cite{li2018benchmarking}, the MLDCP model can optimize the rough estimation of transmission map $t(x)$ and atmospheric light $A$ by self-learning. The authors show experimentally on RESIDE test dataset that their model achieves the highest SSIM and PSNR values (two important full-reference metrics) compared with DCP and some other well-known state-of-the-art dehazing algorithms and CNN-based architectures. The authors further evaluate the effect on object detection in the hazy environment when dehazing the test images by their MLDCP model. Besides, the experimental results demonstrate that MLDCP not only enhances the performance of object detection in the hazy environment but also outperforms most dehazing algorithms on this task with higher detection accuracy.  

The authors also present a straightforward and flexible algorithm to generate synthetic haze to any existing image datasets, inspired by a reversed MLDCP model. The authors aim to enhance object detection performance in the hazy environment by utilizing synthetic hazy images as training datasets. In the experiment, this algorithm is applied to MS COCO training dataset \cite{lin2014microsoft} by adding synthetic haze to the images and build a new Hazy-COCO training dataset. The authors evaluate the mean average precision (mAP) of Mask R-CNN \cite{he2017mask}, a widely adopted object detection and segmentation model, by training the network with the Hazy-COCO training dataset. The experimental results indicate that it leads to an impressive improvement when preprocessing training datasets with the inverse MLDCP algorithm. 

\section{Related Work}
\subsection{Overview of Dehazing Algorithms}
\subsubsection{Background Knowledge}

In computer vision, the widely used atmospheric scattering model to describe the generation of a hazy image is as follows:
\begin{equation} \label{eq:1}
	\textbf{\textit{I}}(x) = \textbf{\textit{J}}(x)\textit{t}(x) + \textbf{\textit{A}}(1 - \textit{t}(x))
\end{equation}
where $I(x)$ is the observed intensity (hazy image), $J(x)$ is the scene radiance (haze-free image), $t(x)$ is the medium transmission map, and $A$ is the atmospheric light. The first term $J(x)t(x)$ is called attenuation and the second term $A(1-t(x))$ is called airlight \cite{tan2008visibility}. 

The medium transmission map $t(x)$ describes the portion of the light that is not scattered and reaches the camera \cite{he2010single}. When the atmosphere is homogeneous, the transmission matrix $t(x)$ can be defined as:
\begin{equation} \label{eq:2}
	\textit{t}(x) = \textit{e}^{-\beta d(x)}
\end{equation}
where $\beta$ is the scattering coefficient of the atmosphere, and $d(x)$ is the scene depth representing the distance between the object and camera.

Most state-of-the-art single image dehazing algorithms exploit the atmospheric scattering model (\ref{eq:1}) and estimate the transmission matrix $t(x)$ and the atmospheric light $A$ in either physically grounded or data-driven ways. Then the haze-free images $J(x)$ can be recovered by computing the reformulation of (\ref{eq:1}):
\begin{equation} \label{eq:3}
	\textbf{\textit{J}}(x) = \frac{1}{\textit{t}(x)}\textbf{\textit{I}}(x) - \textbf{\textit{A}}\frac{1}{\textit{t}(x)} + \textbf{\textit{A}}
\end{equation}

\subsubsection{Conventional Single Image Dehazing Algorithms}
Haze removal is a challenging task due to the non-linear and dependent depth maps over a global scene in hazy images. Many efforts have been made to tackle this challenge by exploiting natural images priors and depth statistics. Most conventional dehazing algorithms focus on predicting two critical parameters, medium transmission matrix $t(x)$ and global atmospheric light $A$, which are necessary to recover haze-free images via computing (\ref{eq:3}). In \cite{tan2008visibility}, an automated method is proposed based on the observation that the contrast of a haze-free image is higher than that of a hazy image. Furthermore, a Markov Random Fields (MRFs) framework is implemented to estimate the atmospheric light $A$ by maximizing the local contrast of a hazy image. The output results are visually impressive but may not be physically valid. Assuming that the transmission and surface shading is uncorrelated in local areas, \cite{fattal2008single} eliminates the scattered light by locally estimating the optical transmission map of the scene with constant constraints. Despite its compelling results, it may fail in the cases with heavy haze and lead to the inaccurate estimation of color and depth maps.

\cite{he2010single} proposed a widely recognized single image dehazing algorithm called Dark Channel Prior (DCP), which can estimate the transmission map $t(x)$ more reliably. A regular pattern is found that in most non-sky patches of haze-free images, at least one color channel (dark channel) has some pixels whose intensity is very low and even close to zero. Then this pixel-based observation can be formally described by defining the dark channel $J^{dark}$ as:
\begin{equation} \label{eq:4}
	\textbf{\textit{J}}^{dark}(x) = \min_{y\in \Omega (x)}(\min_{c} \textbf{\textit{J}}^{c}(y)) \approx 0 
\end{equation}
where $c$ indicates RGB color channels and $y$ refers to the pixel in a local patch $\Omega (x)$ centered at $x$.
Adding minimum operators to both sides of the transformation of (\ref{eq:1}):
\begin{equation} \label{eq:4.5}
	\min_{y\in \Omega (x)}(\min_{c}\frac{\textbf{\textit{I}}^{c}(y)}{\textbf{\textit{A}}^{c}}) = \widetilde{t}\min_{y\in \Omega (x)}(\min_{c}\frac{\textbf{\textit{J}}^{c}(y)}{\textbf{\textit{A}}^{c}}) + 1 - \widetilde{t}
\end{equation}
Transmission map $\widetilde{t}$ can be put outside of the minimum operators based on the fact that $\widetilde{t}$ is a constant in the patch. 

Since the dark channel of a haze-free image can be approximately taken as 0, the multiplicative term in (\ref{eq:4.5}) can be eliminated by adding (\ref{eq:4}). Then transmission map $\widetilde{t}$ can be predicted by:
\begin{equation} \label{eq:5}
	\widetilde{t} = 1 - \omega\min_{y\in \Omega (x)}(\min_{c} \frac{\textbf{\textit{I}}^{c}(y)}{\textbf{\textit{A}}^{c}})
\end{equation}
The additional parameter $\omega$ is a constant parameter that optionally controls the degree of haze removal. Even in a haze-free image, the haze still exists among distant objects. A small amount of haze will keep the vision perceptual natural with the sense of depth. The dehazing parameter determines how much haze will be removed. 

In case that the recovered scene radiance $J(x)$ is prone to noise when transmission map $t(x)$ is extremely low, DCP restricts $t(x)$ by a lower bound $t_{0}$, which is set to 0.1 in \cite{he2010single}:
\begin{equation} \label{eq:6}
    \widetilde{t} = \max(t(x), t_{0})
\end{equation}

As for the estimation of atmospheric light $A$, \cite{tan2008visibility} defines it as the color of the most haze-opaque regions, which refers to the brightest pixels in a hazy image. However, this assumption only applies when there is no sunlight in local regions. This limitation is optimized in \cite{he2010single} by considering the sunlight and adopting the dark channel to detect the most haze-opaque. DCP picks the top 0.1 percent brightest pixels in the dark channel, among which the pixels with the highest intensity in an input image are selected as the atmospheric light.

\subsubsection{Limitations of DCP}
DCP has some limitations that it may fail to accurately estimate transmission map $t(x)$ and atmospheric light $A$, when object surfaces are essentially analogous to the air light over a local scene without any projected shadows. Although DCP \cite{he2010single} takes sunlight into consideration, the influence of sunlight is still tremendous when there is strong sunlight in the sky region. It will underestimate the transmission map of these objects and overestimate the haze layer. Thus the brightness of the restored image is darker than the real-world haze-free image. The authors compare a group of hazy images and recovered images from DCP, the color distortion in the sky region can be observed obviously in Fig. \ref{fig:The color distortion of DCP in sky region}:
\begin{figure}[H]
    \subfigure[Real-world Hazy image]
    {
        \includegraphics[width=1.6in]{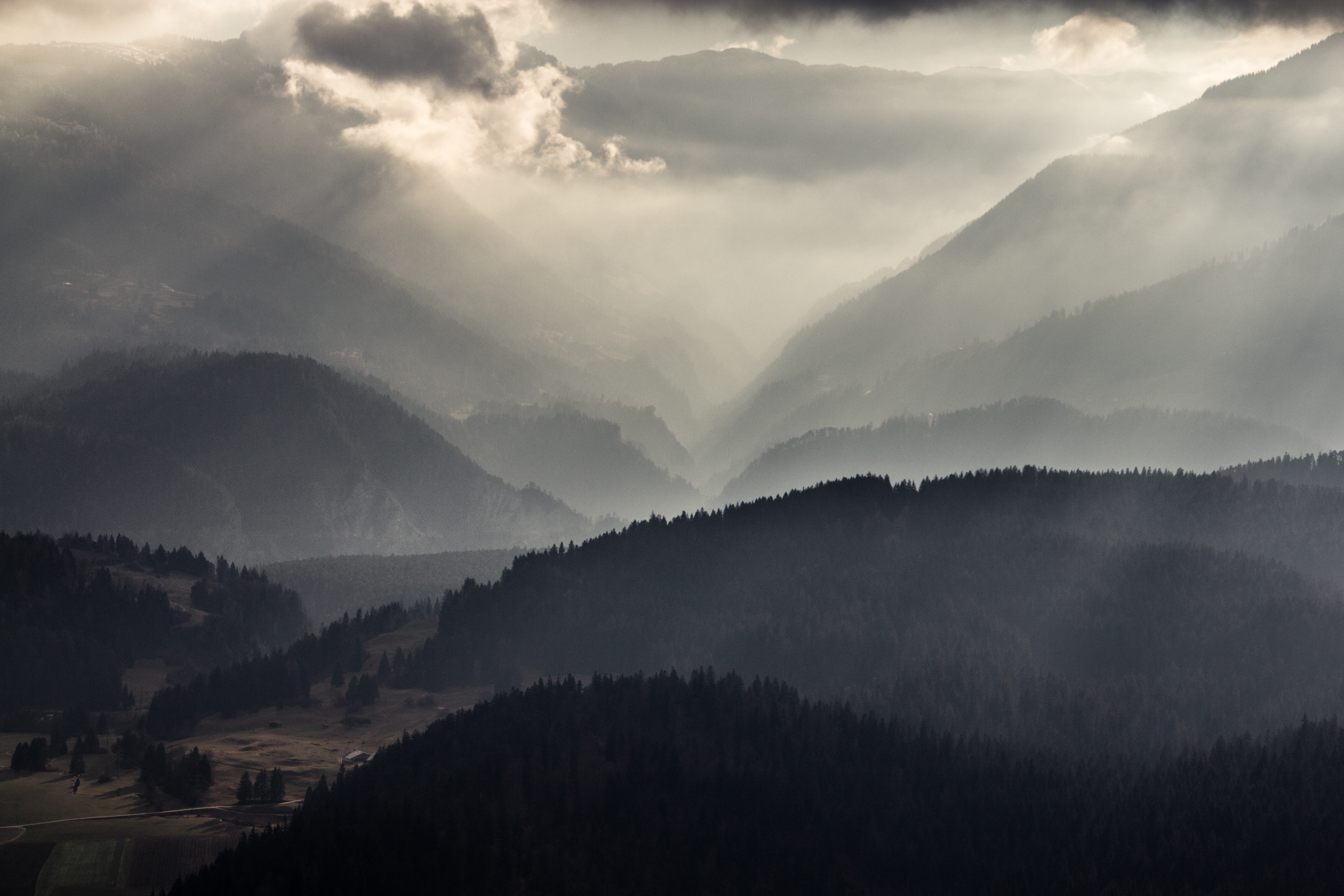}
        \label{b_8}
    }
    \subfigure[Dehazed image via DCP]
    {
        \includegraphics[width=1.6in]{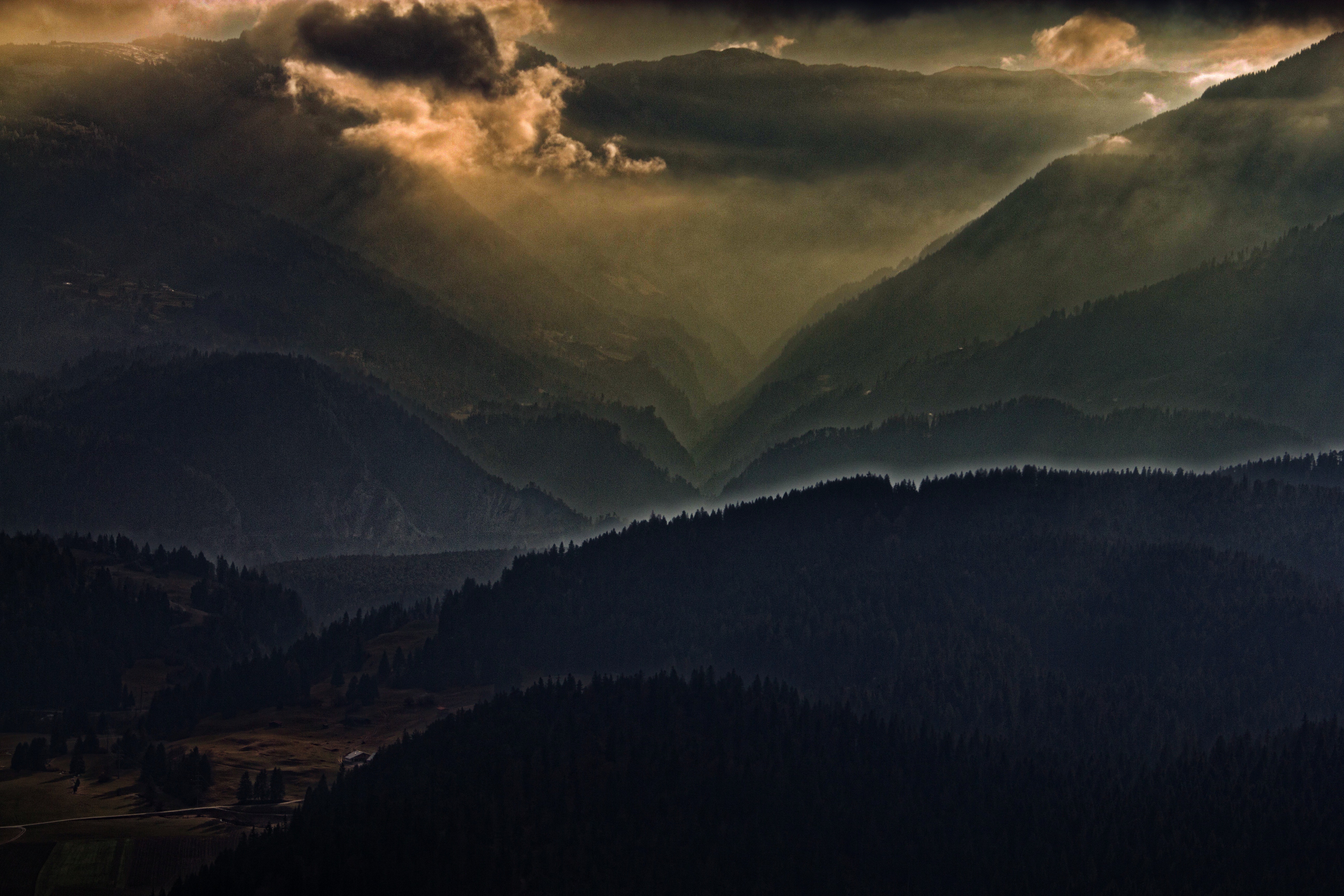}
        \label{b_9}
    }
    \\
    \subfigure[Real-world Hazy image]
    {
        \includegraphics[width=1.6in]{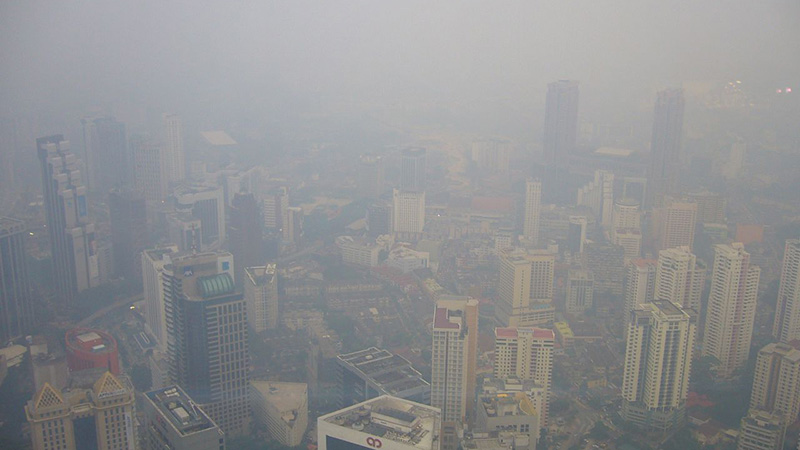}
        \label{b_8}
    }
    \subfigure[Dehazed image via DCP]
    {
        \includegraphics[width=1.6in]{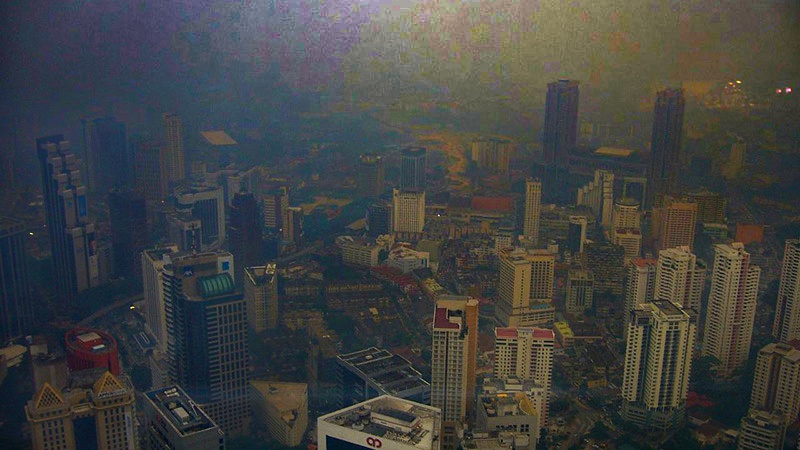}
        \label{b_9}
    }
    \caption{Limitations of DCP}
    \label{fig:The color distortion of DCP in sky region}
\end{figure}
Additionally, the constant parameter $\omega$ in (\ref{eq:5}) is fixed to 0.95 in \cite{he2010single} without any changes corresponding to different haze distributions. Excessive haze removal will create color distortion, and images with insufficient haze removal will remain blurred. In order to solve this problem, \cite{chen2017adaptive} proposes a dehazing parameter adaptive method based on DCP that estimates dehazing parameter $\omega$ locally instead of globally. It can automatically adjust the value of $\omega$ according to the distribution of haze. DCP in \cite{he2010single} implements soft matting to optimize the estimation of transmission map $t(x)$, which is further enhanced to be more accurate and efficient by utilizing an explicit image filter called $guided filter$ in \cite{he2012guided}. \cite{zhu2017single} points out that traditional DCP has not fully exploited its potential and will generate undesirable artifacts due to inappropriate assumptions or operations. Then it introduces a novel method that estimates transmission map $t(x)$ by energy minimization. The energy function combines DCP with piecewise smoothness and obtains an outstanding performance compared to conventional pixel-based dehazing algorithms. Several attempts have been made in \cite{kim2013optimized} \cite{shen2013sky} \cite{shi2016image} to process the color distortion and optimize the restoration in some bright regions, such as the sky and reflective surfaces.

\subsubsection{Overview of CNN-based Dehazing Algorithms}
In recent years, neural network has made significant progress in numerous computer vision tasks \cite{hua2019dilated} \cite{zhang2019bend} \cite{wei2020deepsfm} and natural language processing tasks \cite{wang2020attention} \cite{wang2017comparisons} \cite{wang2017combining}. Various Convolutional Neural Networks (CNNs) are designed to obtain a more accurate estimation of transmission matrix $t(x)$ by self-learning the mapping between hazy images and corresponding transmission maps, which outperform most conventional dehazing algorithms. DehazeNet in \cite{cai2016dehazenet} is an end-to-end system built upon a deep convolutional neural network whose layers are specially designed to embody established priors in haze removal. Furthermore, a novel non-linear activation function is executed to improve the quality of output recovered images. \cite{ren2016single} proposes a multi-scale deep neural network to dehaze a single image, which consists of a coarse-scale stage that roughly predicts the transmission matrix $t(x)$ over a global view, and a fine-scale stage that refines the rough estimation locally. \cite{golts2019unsupervised} utilize the DCP energy function as the loss function in a fully-convolutional dilated residual network. Feeding the network with real-world outdoor images, it minimizes the loss function completely unsupervised during the training process. A light-weight CNN model called All-in-One Dehazing Network (AOD-Net) \cite{li2017all} is designed based on a reformulated atmospheric scattering model. It generates the recovered images directly and can be widely embedded into other deep CNN models to enhance the performances of some high-level tasks over hazy images.

\subsubsection{Dehazing Benchmark Datasets and Metrics}
Traditional haze-removal algorithms used to evaluate and compare dehazing performances by merely presenting a group of hazy images and dehazed images restored from various dehazing algorithms. The enhancement of dehazing performance is expected to be observed from the visual comparison of images. However, it is not convincing to prove that a new dehazing algorithm outperforms other algorithms only from human eyes perception. Two widely adopted image metrics to evaluate and compare single image dehazing algorithms are PSNR and SSIM \cite{hore2010image}. PSNR refers to the peak-signal-to-noise ratio, which is generally applied to evaluate the image quality. SSIM refers to the structural similarity index measure, a well-known metric to measure the similarity between two images. Since it is generally impossible to capture the same visual scene with and without haze, while all other environmental conditions stay identical, it is incredibly challenging to measure the SSIM value between a hazy image and its haze-free ground truth image. Therefore, recent efforts have been made to create synthetic hazy images from haze-free images based on the depth information.

\cite{tarel2012vision} builds two sets of images without haze and with synthetic haze from both real-world camera captured scenes and synthetic scenes to evaluate the performance of the proposed visibility enhancement algorithm. They utilize the software to generate 66 synthetic images built upon a physically-based road environment. By obtaining the depth information, four different fog types are added to 10 camera images and finally create a dataset with over 400 synthetic images in total. A fog simulation is proposed in \cite{sakaridis2018semantic} by simulating the underlying mechanism of hazy image formation (\ref{eq:1}) and utilizing the standard optical model for daytime haze. The fog simulation pipeline is leveraged to add synthetic fog to urban scenes images in Cityscapes dataset \cite{cordts2016cityscapes} and generate a Foggy Cityscapes dataset. Foggy Cityscapes dataset consists of 550 refined high-quality synthetic foggy images with detailed semantic annotations and additional 20000 synthetic foggy images without sufficient annotations. 

The REalistic Single Image DEhazing (RESIDE) dataset \cite{li2018benchmarking} is the first large-scale dataset for benchmarking single image dehazing algorithms, and it includes both indoor and outdoor hazy images. RESIDE dataset also contains a large-scale synthetic training set and two sets designed respectively for objective and subjective quality evaluations. Moreover, in the supplementary RESIDE-$\beta$ set, they add annotations and object bounding boxes to an evaluation set consisting of 4322 real-world hazy images, which can be utilized to test the performance of object detection in the hazy environment. \cite{li2018benchmarking} also provides a rich variety of criteria beyond PSNR and SSIM to evaluate the performance of dehazing algorithms, including full-reference metrics, no-reference metrics, subjective evaluation, and task-driven evaluation. However, the criteria are only practically applicable to the global performance of haze-removal. It cannot embody the difference locally between two images, failing to judge if our MLDCP model outperforms traditional DCP specifically in the bright region. Therefore, the authors compare dehazing performances experimentally with both recovered images' visual quality and two pivotal metrics PSNR and SSIM. 

\subsection{Overview of Object Detection Models}
Object detection is the combination of object classification and localization, which can both recognize and localize all object instances of specific categories in an image. Due to its close relationship with image and video analysis, object detection has been widely applied in various computer vision tasks, especially in autonomous driving, traffic surveillance, and some other smart city fields. 

Fast R-CNN refers to Fast Region-based Convolutional Network \cite{girshick2015fast}. The network is fed with an image and a set of object proposals and outputs a convolutional feature map. A region of interest (RoI) pooling layer is proposed to extract a fixed-length feature vector and feed it into a sequence of fully connected layers. The output layers contain a softmax layer that estimates the softmax probability over $K$ object classes plus a background class, and another layer with offset values that refine the bounding box positions of an object. Fast R-CNN overcomes the disadvantages of R-CNN and SPPnet \cite{he2015spatial} with a higher detection accuracy as well as faster training and test speed. Later in \cite{ren2015faster}, a Region Proposal Network (RPN) is added into Fast R-CNN generating a Faster R-CNN. The RPN aims to simultaneously propose candidate object bounding boxes and corresponding scores at each position. It then implements RoIPool in Fast R-CNN to extract features from each candidate box and perform classification as well as bounding-box regression. Both RPN and Fast R-CNN are trained independently but share the same convolutional layers. While achieving state-of-the-art object detection accuracy, the unified network can further increase the speed significantly. 

Mask R-CNN \cite{he2017mask} extends the Faster R-CNN by an output branch with a binary mask for each RoI, in parallel to a branch for bounding-box recognition and classification. However, the quantization of RoIPool in Faster R-CNN has a negative effect on predicting binary masks. To remove this harsh quantization, Mask R-CNN replaces RoIPool with a RoIAligh layer that aligns the extracted features with the input properly. It can efficiently detect object instances in an image and simultaneously generate a pixel-accurate segmentation mask for each instance. Due to its compelling performance and influential architecture, Mask R-CNN is widely used as a solid baseline to exploit more object detection tasks.

\subsection{Domain Adaption Methods}
Domain adaptation is a novel strategy that can be utilized to advance object detection models \cite{patel2015visual} \cite{li2013learning} \cite{gopalan2011domain}. And it has been proved effective, especially in some extreme weather \cite{chen2018domain} \cite{zhe2018domain}, such as hazy, rainy, and snowy. \cite{chen2018domain} has proposed a Domain Adaptive Faster R-CNN to enhance the cross-domain robustness of object detection. Based on H-divergence theory, two domain adaptation components on image level and instance level are integrated into Faster R-CNN architecture, aiming to reduce the domain discrepancy at both levels. Training data with images and full supervision is used as the source domain, and only unlabeled images in test data are available for the target domain. For both components, it adapts the classifier trained on a source domain and implements the adversarial training strategy to learn domain-invariant features. The method further incorporates a consistency regularization into the Faster R-CNN model to obtain a domain-invariant region proposal network (RPN).

Inspired by Domain Adaptive Faster R-CNN in \cite{chen2018domain}, \cite{liu2018improved} adopts a similar idea and designs a Domain-Adaptive Mask-RCNN (DMask-RCNN). The source domain takes the clean images in the MS COCO dataset, and the target domain takes unannotated real-world hazy images in RESIDE dataset \cite{li2018benchmarking}, and their dehazed output images by MSCNN \cite{ren2016single} respectively. And  DMask-RCNN adds a domain-adaptive branch after the base feature extraction layers in Mask R-CNN architecture, aiming to mask the generated features to be domain-invariant between the source domain and target domain. The experimental results in \cite{chen2018domain} \cite{liu2018improved} demonstrate that the domain adaptation method can enhance the performance of both Faster R-CNN and Mask R-CNN models when tackling the object detection task in the hazy environment. Moreover, this enhancement can be more effective when feeding the target domain with images restored by a robust dehazing algorithm.

\subsection{Object Detection Datasets}
Object recognition is a core task to understand a visual scene in computer vision, which involves several sub-tasks, including image classification, object detection, and semantic segmentation. All three tasks have high demands for image datasets. Object classification requires each image to be labeled with a sequence of binary numbers, which indicate if object instances exist in the image or not. Object detection is more challenging, which combines object classification and localization. It not only identifies which specific class an object belongs to but also locates it in an image. The object localization requires collecting bounding boxes that locate one or more objects in an image, which is a considerable workload in a large-scale image dataset. The PASCAL VOC challenge \cite{everingham2010pascal} is a widely recognized benchmark in visual object recognition and detection. The VOC2007 dataset contains 20 object categories spread over 11000 images. The annotation procedure is designed to be consistent, accurate, and exhaustive, and annotations are made available for training, validation, and test data. The ImageNet dataset \cite{deng2009imagenet} involves millions of cleanly labeled and full-resolution images in the hierarchical structure of WordNet \cite{miller1998wordnet}. The dataset provides an average of 500-1000 images to illustrate each synset in a majority of 80000 synsets of WordNet. Many object recognition algorithms have made a significant breakthrough by using the training resource of the ImageNet dataset. This benchmark dataset also plays an essential role in advancing object recognition tasks. 

Semantic segmentation requires labeling each pixel in an image to a category, and it is an extremely time-consuming task to build a large-scale dataset with detailed semantic scene labeling. The Microsoft COCO dataset \cite{lin2014microsoft} consists of about 3.3 million images, and over 2 million of them are labeled. It collects 91 common object categories, which are fewer than the ImageNet dataset. However, in contrast to both the VOC2007 dataset and ImageNet dataset, MS COCO dataset contains significantly more instances per category and considerably more object instances per image. And every instance of each object category is fully labeled and segmented by a novel instance-level segmentation mask. MS COCO dataset has been widely utilized for training some more complex CNN architectures that aim to make further progress on object recognition as well as semantic segmentation tasks.  

\section{Proposed Methods}
\subsection{Multiple Linear Regression DCP Model}
DCP estimates transmission map $t(x)$ from (\ref{eq:5}) and selects the pixels in the dark channel with the highest intensity among the top 0.1 percent brightest pixels as the atmospheric light $A$. Then hazy images can be recovered from (\ref{eq:3}) motivated by the estimations of $t(x)$ and $A$. However, the rough estimations in DCP introduced in Section 2.1.3 can generate some unpredicted deviations, which are theoretically impossible to be eliminated during the estimation process. Now the authors implement a multiple linear regression model to optimize the haze-removal algorithm in DCP.

Multiple linear regression is a statistical technique using several explanatory variables to predict the output response variable. As a predictive analysis, multiple linear regression aims to model the linear relationship between two or more independent variables and a continuous dependent variable. The authors still adopt the rough estimations on transmission map $t(x)$ and atmospheric light $A$ in DCP. Three components $\frac{\textbf{\textit{I}}(x)}{\textit{t}(x)}$, $\frac{\textit{A}}{\textit{t}(x)}$ and $\textit{A}$ in (\ref{eq:3}) can be regarded as explanatory variables, since $\textbf{\textit{I}}(x)$ refers to the pixel of input hazy image while $t(x)$ and $A$ can be estimated from DCP. The scene radiance $\textbf{\textit{J}}(x)$, which refers to the pixel of output recovered image, can be regarded as the response variable. The authors add regression coefficient weights to each explanatory variable and a constant term (bias) to the atmospheric scattering model. Then (\ref{eq:3}) can be reformulated as a multiple linear regression model (\ref{eq:7}), which describes how the mean response $\textbf{\textit{J}}(x)$ changes with explanatory variables:

\begin{equation} \label{eq:7}
\textbf{\textit{J}}(x) = \omega_{0} \frac{\textbf{\textit{I}}(x)}{\textit{t}(x)} + \omega_{1} \frac{\textbf{\textit{A}}}{\textit{t}(x)} + \omega_{2} \textbf{\textit{A}} + b
\end{equation}

As the authors introduced, DCP is already an effective dehazing algorithm even with the rough estimations on $t(x)$ and $A$. The authors implement a multiple linear regression model to optimize DCP by learning the relationships between hazy images with haze-free images, which is extremely challenging to find out in traditional pixel-based dehazing algorithms. Since both $\textbf{\textit{J}}(x)$ and $A$ are defined and estimated on RGB color channels, the dimensions of three weights and a bias should be (3,1), intending to refine the parameters at all three color channels. The Outdoor Training Set (OTS) in RESIDE dataset \cite{li2018benchmarking} provides 8970 outdoor haze-free images, each of which also contains 30 synthetic hazy images with haze intensity ranging from low to high. During the training process, the network learns the relationships between output recovered images and their relevant haze-free ground truth images. However, it is theoretically impossible to obtain the ground truth image of a real-world hazy image. The most efficient approach is referring to a real-world haze-free image as ground truth image and its synthetic hazy image as input image respectively. The various intensity of synthetic haze guarantees that the MLDCP model can be applied to more than a fixed haze intensity in real-world scenes.

When MLDCP model is trained on OTS dataset, the authors refer to haze-free images as target ground truth images $\textbf{\textit{J}}$, refer to synthetic hazy images as input images $\textbf{\textit{I}}$, and refer to the recovered images as output images $\textbf{\textit{J}}_{\omega}$. In order to simplify the formula (\ref{eq:7}) during the training process, the authors assign $\textbf{\textit{x}}_{0} = \frac{\textbf{\textit{I}}(x)}{\textit{t}(x)}$, $\textbf{\textit{x}}_{1} = \frac{\textit{A}}{\textit{t}(x)}$ and $\textbf{\textit{x}}_{1} = \textit{A}$. Then (\ref{eq:7}) can be reformulated as:
\begin{equation} \label{eq:8}
    \textbf{\textit{J}}_{\omega}(x) = \omega_{0} \textbf{\textit{x}}_{0} + \omega_{1} \textbf{\textit{x}}_{1} + \omega_{2} \textbf{\textit{x}}_{2} + b
\end{equation}

The deviations between the output images $\textbf{\textit{J}}_{\omega}$ and the target ground truth images $\textbf{\textit{J}}$ are estimated by mean-squared error (MSE):
\begin{equation} \label{eq:9}
    \textbf{\textit{MSE}} = (\textbf{\textit{J}} - \textbf{\textit{J}}_{\omega})^{2}
\end{equation}

MSE measures the average squared difference between an observation’s actual and predicted values, which is commonly adopted in linear regression models. The loss function of MLDCP is evaluated based upon MSE:
\begin{equation} \label{eq:10}
    \textbf{\textit{L}}(\omega) = \frac{1}{2n} \sum_{i=1}^{n} (\textbf{\textit{J}}^{(i)} - \textbf{\textit{J}}_{\omega}^{(i)})^{2}
\end{equation}

During the training process, the authors calculate the gradient of loss function by Stochastic Gradient Descent (SGD), a widespread and effective technique to address optimization tasks. Then the optimal weights and bias can be obtained by minimizing MSE through the following steps: 

\begin{itemize}
\item Implement SGD to compute the derivative of loss function to find the gradient of the error generated from current weights and bias, and adjusts the weights by moving in the direction opposite of the gradient to decrease MSE. Then authors iteratively update three weights via:
\begin{equation} \label{eq:11}
    \omega_{k} = \omega_{k} - \alpha \frac{\partial }{\partial \omega} \textbf{\textit{L}}(\omega)
\end{equation}
where $\alpha$ is the learning rate, and $k \in {(0,1,2)}$. 

\item Expand the derivative of loss function (\ref{eq:10}), then (\ref{eq:11}) can be reformulated as:
\begin{equation} \label{eq:12}
    \omega_{k} = \omega_{k} - \alpha \frac{1}{n} \sum_{i=1}^{n} (\textbf{\textit{J}}^{(i)} - \textbf{\textit{J}}_{\omega}^{(i)})x_{k}
\end{equation}
\item Update the bias with the same process repeatedly by training each image in OTS dataset via:
\begin{equation} \label{eq:13}
    b = b - \alpha \frac{1}{n} \sum_{i=1}^{n} (\textbf{\textit{J}}^{(i)} - \textbf{\textit{J}}_{\omega}^{(i)})
\end{equation}
\end{itemize}

When the optimal weights and bias are obtained from the training process, they can be added to the haze-removal model (\ref{eq:7}) directly, and real-world hazy images can be dehazed in the same way as that in conventional DCP. The improved performance is determined by the quality of synthetic hazy images. If the synthetic haze is more close to the real-world haze, the MLDCP model can learn more realistic haze information during the training process, and the dehazing performance can be further enhanced. 

\subsection{Synthetic Haze Generated Model}
There have been some large-scale image datasets released for the object detection task. However, few image datasets are built just for object detection in extreme weather like hazy and rainy. The Real-world Task-driven Testing Set (RTTS) in RESIDE \cite{li2018benchmarking} contains 4322 real-world hazy images annotated with both object categories and bounding boxes, which is an essential contribution taken as a test dataset for object detection task in the hazy environment. However, even with data augmentation techniques, its amount of training images is far less efficient in training an object detection model in comparison with the larger scale of the MS COCO training dataset \cite{lin2014microsoft}. It means a considerable workload to build a large-scale image dataset with detailed annotations and segmentation instances like MS COCO or PASCAL VOC. And the rarity of real-world hazy images, especially in urban cities, makes it more challenging to collect the same amount of high-quality hazy images as those in the MS COCO training dataset.

The authors propose an algorithm that generates synthetic haze to any existing large-scale haze-free image datasets without much computation. Inspired by the multiple linear regression haze-removal model, the authors propose a new algorithm that implements a similar inverse MLDCP model to add synthetic haze to an image. If an object detection model is trained on the synthetic hazy dataset generated by this algorithm, its performance of detecting an object in hazy weather can be significantly enhanced.

Since DCP is an effective dehazing algorithm, and the multiple linear regression model can further exploit its potential, the authors suppose that similar ideas can also be applied to synthesize higher quality hazy images based on the regular dark channel pattern. Coefficient weights and bias are added to (\ref{eq:1}), which is expected to maximally restore the deviations between haze-free images and their relevant synthetic hazy images. Then the reformulated atmospheric scattering model is obtained as follows:
\begin{equation} \label{eq:14}
    \textbf{\textit{I}}(x) = \beta_{0}\textbf{\textit{J}}(x)\textit{t}(x) + \beta_{1}\textbf{\textit{A}}\textit{t}(x) + \beta_{2}\textbf{\textit{A}} + b
\end{equation}
The synthetic haze generated model is still trained on OTS in RESIDE dataset \cite{li2018benchmarking}. In contrast to MLDCP model, the authors refer to haze-free images as input images $\textbf{\textit{J}}$ and refer to synthetic hazy images in OTS as target images $\textbf{\textit{I}}$. And the output synthetic hazy images are taken as $\textbf{\textit{I}}_{\beta}$. The cost function is defined based on MSE that evaluates the deviations between the output images $\textbf{\textit{I}}_{\beta}$ and target images $\textbf{\textit{I}}$:
\begin{equation} \label{eq:15}
    \textbf{\textit{L}}(\beta) = \frac{1}{2n} \sum_{i=1}^{n} (\textbf{\textit{I}}^{(i)} - \textbf{\textit{I}}_{\beta}^{(i)})^{2}
\end{equation}

The authors implement SGD during the training process to travel down the slope of cost function until it reaches the bottom lowest value, and three weights and a bias are updated iteratively by calculating the derivative of cost function as follows:
\begin{equation} \label{eq:16}
    \beta_{k} = \beta_{k} - \alpha \frac{1}{n} \sum_{i=1}^{n} (\textbf{\textit{I}}^{(i)} - \textbf{\textit{I}}_{\beta}^{(i)})x_{k}
\end{equation}

\begin{equation} \label{eq:17}
    b = b - \alpha \frac{1}{n} \sum_{i=1}^{n} (\textbf{\textit{I}}^{(i)} - \textbf{\textit{I}}_{\beta}^{(i)})
\end{equation}
where $k \in {(0,1,2)}$ and $x_{k}$ equals to $\textbf{\textit{J}(x)}\textit{t}(x)$, $\textit{A}\textit{t}(x)$, and $\textit{A}$ in (\ref{eq:14}) respectively. 

In fact, traditional DCP does not support the inverse functionality of generating synthetic haze to a haze-free image. Self-learning the haze information from input synthetic hazy images, the MLDCP model fully utilizes the superiority of DCP to generate a much higher quality synthetic haze with a more similar visual and inherent effect to real-world haze. Additionally, the inverse MLDCP model can add various intensities of haze to an image dataset, making it possible to train an object detection model on training datasets with different synthetic haze intensities corresponding to the density of real-world haze in test images. Since the method focuses on the enhancement by preprocessing the training dataset, it can be applied simultaneously with any improvements on the object detection models or preprocessing images in test datasets with more effective dehazing algorithms.

\section{Experiment Results and Analysis}
\subsection{MLDCP Dehazing Performance}
\subsubsection{Experiment Setup}
In the experiment, both training and test datasets are obtained from RESIDE dataset \cite{li2018benchmarking}. In the training process, 8970 haze-free images and their corresponding synthetic hazy images in the OTS dataset are utilized for training the MLDCP model. In the test phase, the Synthesis Object Testing Set (SOTS) is used to test the dehazing performance. Since MLDCP focuses on recovering outdoor hazy scenes, the authors do not evaluate its dehazing performance over the 500 synthetic indoor images in SOTS. The dehazing performance of MLDCP is compared with conventional DCP on both SSIM and PSNR metrics over 500 outdoor synthetic hazy images and their haze-free ground truth images. 

\subsubsection{Dehazing Performance on SOTS Dataset}

\begin{figure}[H]
    \subfigure[Synthetic hazy image]
    {
        \includegraphics[width=1.7in]{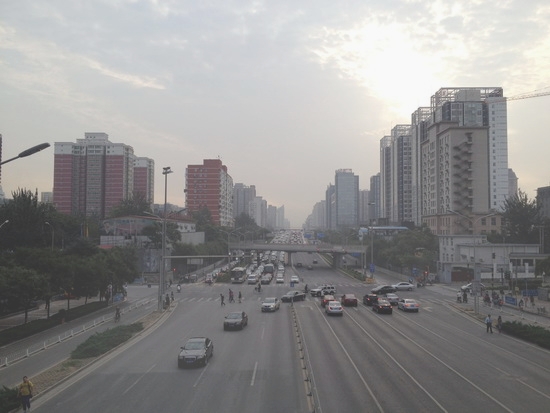}
        \label{fig2_11}
    }
    \subfigure[Haze-free image]
    {
        \includegraphics[width=1.7in]{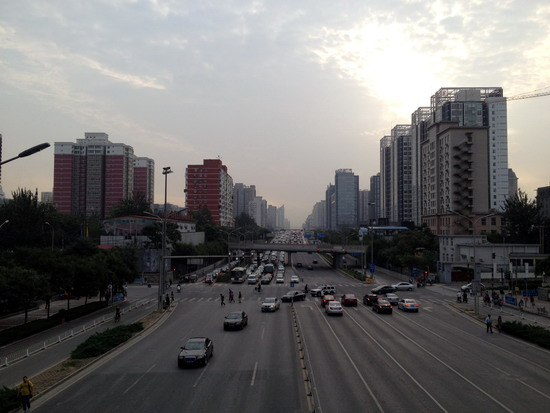}
        \label{fig2_12}
    }
    \\
    \subfigure[Recovered image by DCP]
    {
        \includegraphics[width=1.7in]{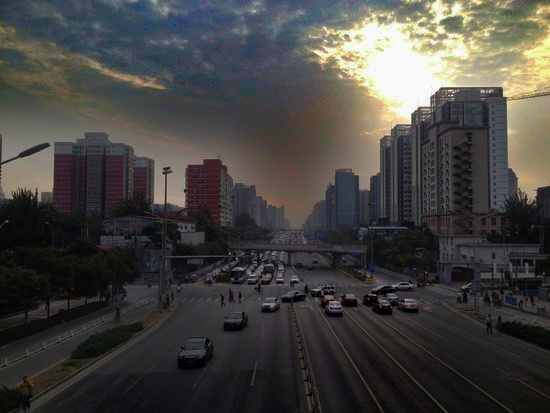}
        \label{fig2_21}
    }
    \subfigure[Recovered image by our model]
    {
        \includegraphics[width=1.7in]{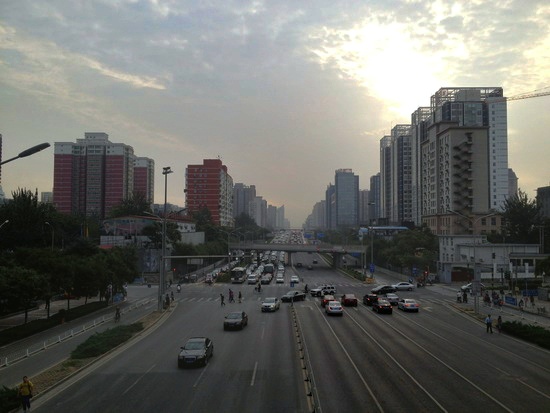}
        \label{fig2_22}
    }
    \caption{Comparison on synthetic hazy image. (a) is a synthetic hazy image in SOTS, (b) is the haze-free ground truth image, (c) is recovered by conventional DCP algorithm, (d) is recovered by our MLDCP model.}
    \label{fig:Comparison on recovered image}
\end{figure}

In Fig. \ref{fig:Comparison on recovered image}, the authors present the comparison of recovered images by conventional DCP and our MLDCP model. It is obviously observed that the recovered image by DCP (c) suffers from the color distortion in the sky region. In contrast, the recovered image by MLDCP (d) performs competently even in the sky region with strong sunlight. And the non-sky region in (c) is much darker than that in (d), and it looks unreal. In comparison with the input synthetic hazy image (a), MLDCP almost removes the haze completely. Besides, its recovered image (d) looks natural and shares a high similarity with the ground truth haze-free image (b).

\begin{table}[H]
    \centering
    \caption{\footnotesize{Average SSIM and PSNR comparison among different dehazing algorithms over 500 outdoor synthetic hazy images in SOTS.}}
    \begin{tabular}{c|c|c}
    \hline
    \multicolumn{3}{c}{\textbf{500 outdoor images}}   \\ 
    \hline
    \textbf{Dehazing method name}  & \textbf{PSNR}    & \textbf{SSIM} \\
    \hline
    \textbf{Improved DCP model}  &   23.84  &  \textbf{0.9411} \\
    \hline
    \textbf{DCP}  &	 18.54	&	0.7100 \\
    \hline
    \textbf{FVR}  &   16.61  & 0.7236 \\
    \hline
    \textbf{CAP}  &  23.95   &  0.8692 \\
    \hline
    \textbf{NLD}  &  19.52   &  0.7328 \\
    \hline
    \textbf{BCCR}  &   17.71  &  0.7409 \\
    \hline
    \textbf{GRM}  &  20.77   &  0.7617 \\
    \hline
    \textbf{DehazeNet}  &   \textbf{26.84}  & 0.8264 \\
    \hline
    \textbf{MSCNN}  &   21.73  & 0.8313 \\
    \hline
    \textbf{AOD-Net}  &  24.08   & 0.8726 \\
    \hline
    \end{tabular}
    \label{tb:outdoor_results}
\end{table}

The authors compare the dehazing performance of MLDCP with several conventional dehazing algorithms as well as some CNN-based haze removal models over two major image benchmark metrics PSNR and SSIM. PSNR evaluates the quality of recovered images, while SSIM measures the similarity between recovered images and haze-free ground truth images. In Table \ref{tb:outdoor_results}, MLDCP model increases PSNR and SSIM by 5.3 and 0.23 respectively in comparison with the conventional DCP. Compared with the performances of other widely adopted dehazing algorithms obtained from \cite{li2018benchmarking}, the proposed MLDCP model achieves a reasonably acceptable PSNR value and the highest SSIM value. Considering that SSIM is essentially more dominant than PSNR when indicating the pixel-wise effect of haze removal, MLDCP outperforms all other dehazing algorithms in Table \ref{tb:outdoor_results}.

\subsubsection{Real-world Hazy Images Dehazing}

\begin{figure}[H]
    \subfigure
    {
        \includegraphics[width=1.1in]{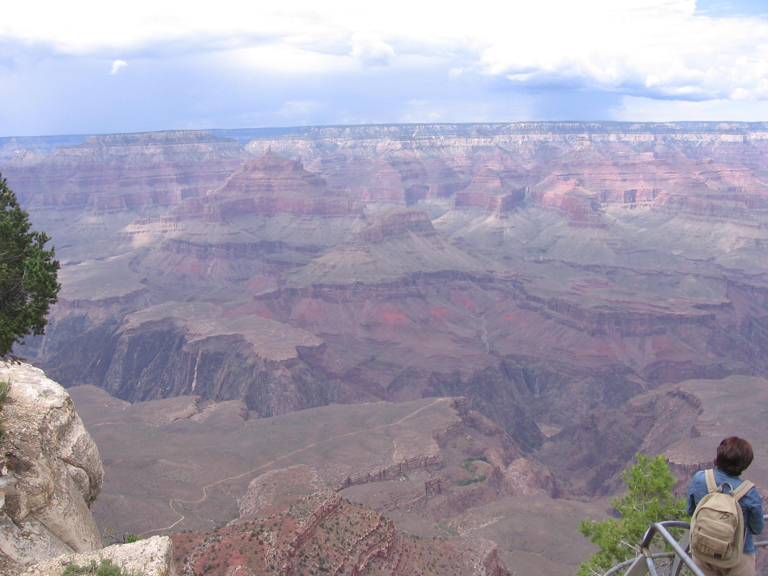}
        \label{A_1}
    }
    \subfigure
    {
        \includegraphics[width=1.1in]{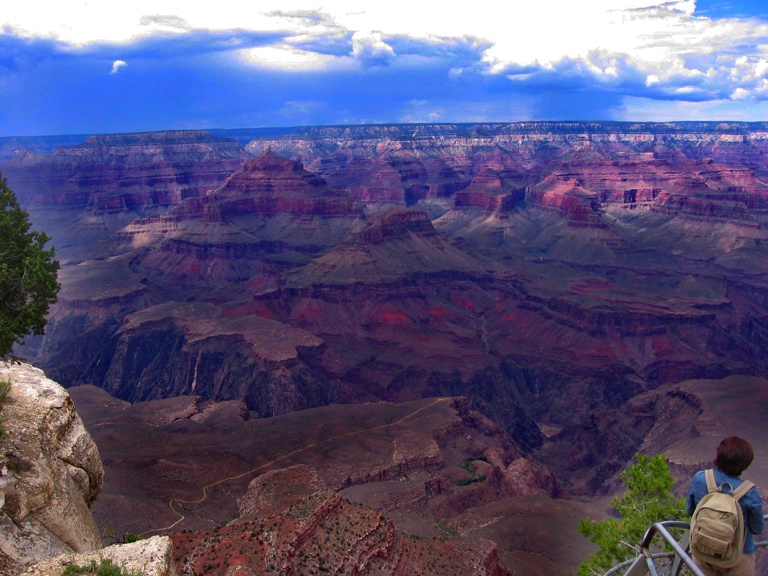}
        \label{A_2}
    }
    \subfigure
    {
        \includegraphics[width=1.1in]{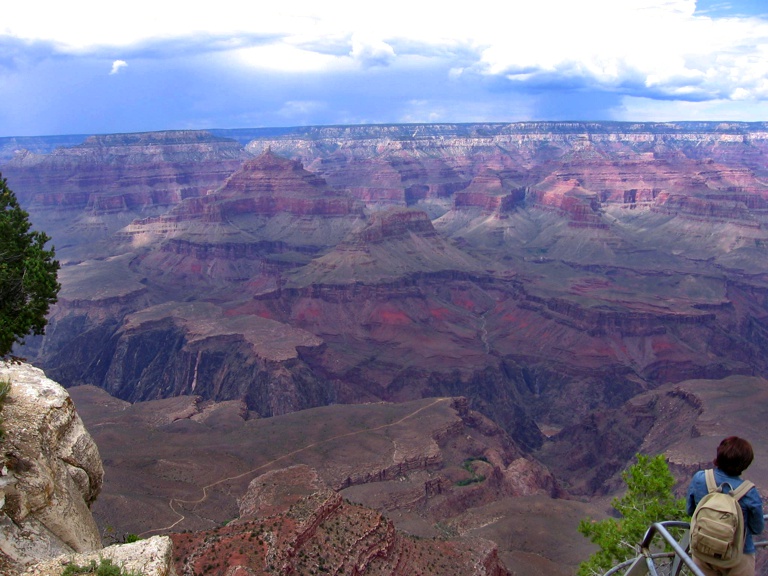}
        \label{A_3}
    }
    \\
    \subfigure
    {
        \includegraphics[width=1.1in]{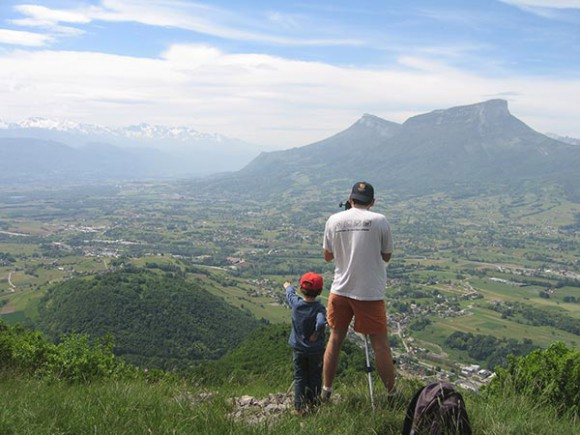}
        \label{B_1}
    }
    \subfigure
    {
        \includegraphics[width=1.1in]{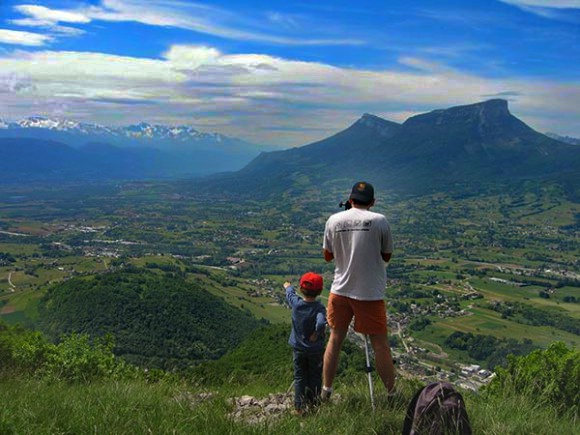}
        \label{B_2}
    }
    \subfigure
    {
        \includegraphics[width=1.1in]{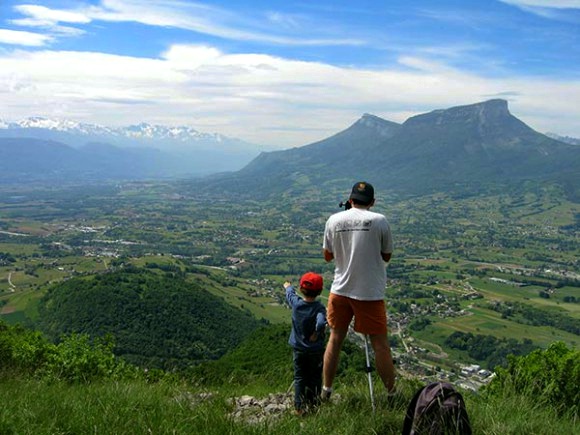}
        \label{B_3}
    }
    \\
    \subfigure
    {
        \includegraphics[width=1.1in]{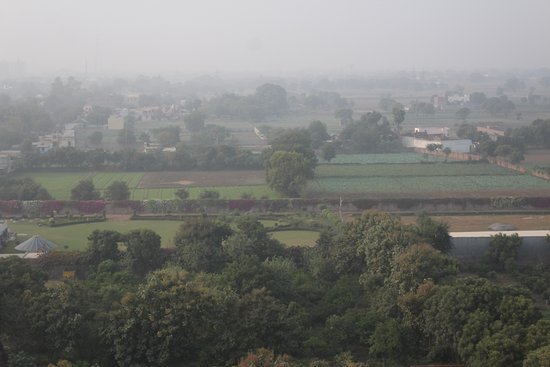}
        \label{C_1}
    }
    \subfigure
    {
        \includegraphics[width=1.1in]{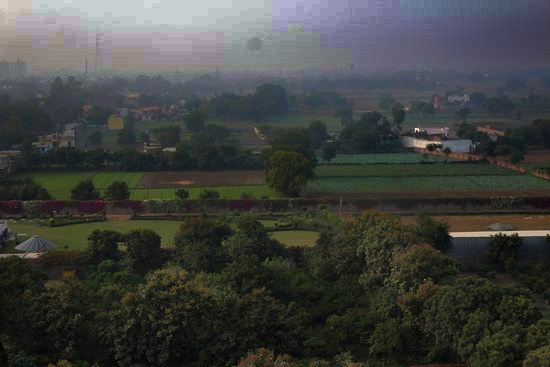}
        \label{C_2}
    }
    \subfigure
    {
        \includegraphics[width=1.1in]{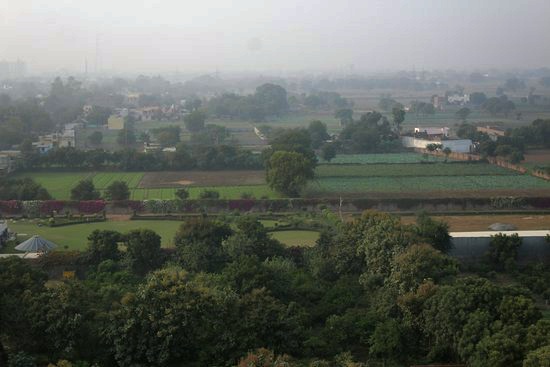}
        \label{C_3}
    }
    \\
    \subfigure
    {
        \includegraphics[width=1.1in]{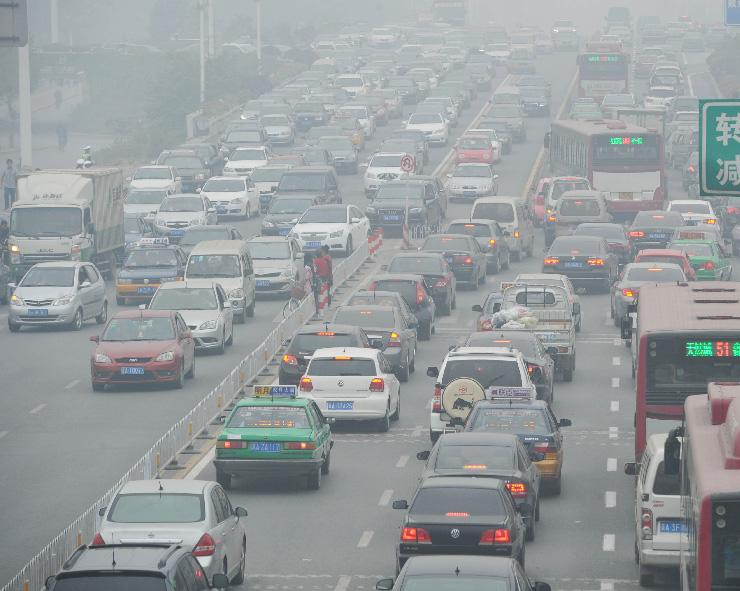}
        \label{D_1}
    }
    \subfigure
    {
        \includegraphics[width=1.1in]{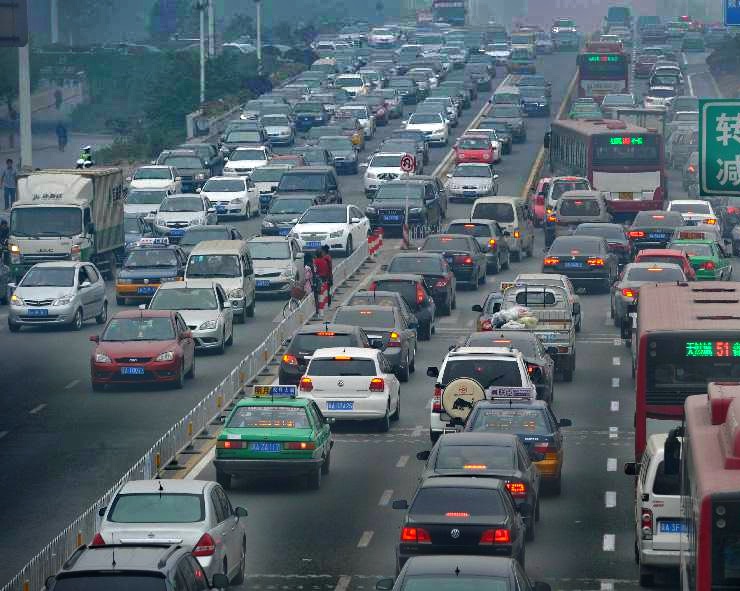}
        \label{D_2}
    }
    \subfigure
    {
        \includegraphics[width=1.1in]{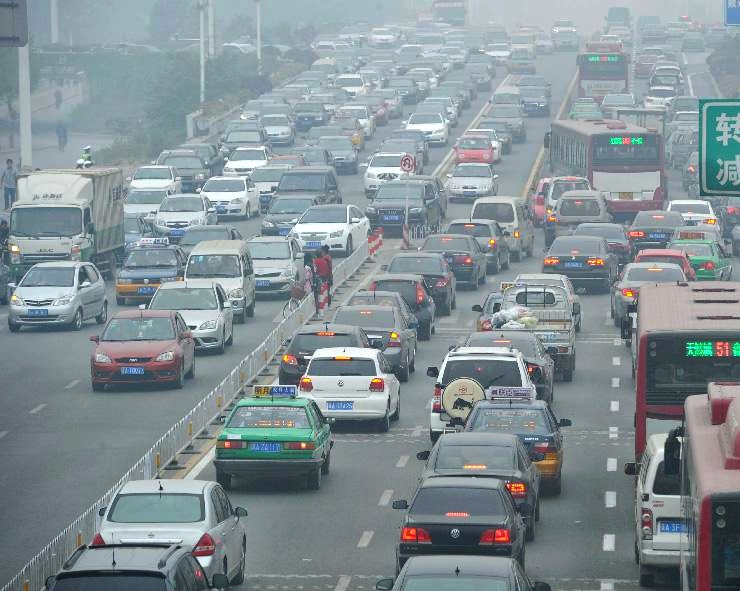}
        \label{D_3}
    }
    \\
        \subfigure
    {
        \includegraphics[width=1.1in]{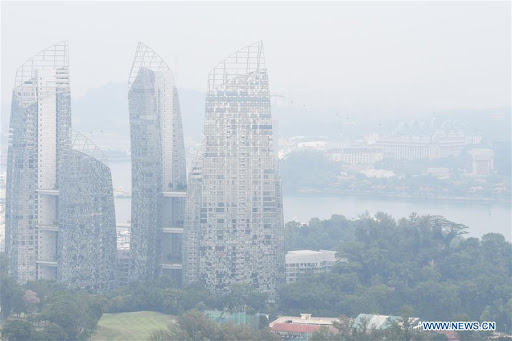}
        \label{E_1}
    }
    \subfigure
    {
        \includegraphics[width=1.1in]{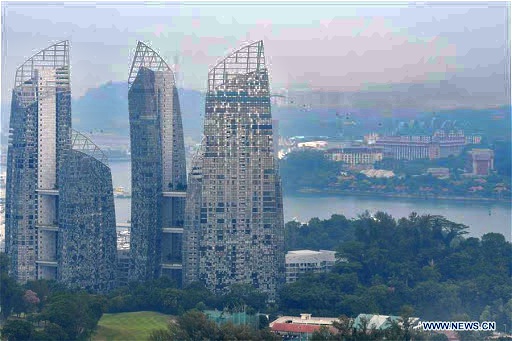}
        \label{E_2}
    }
    \subfigure
    {
        \includegraphics[width=1.1in]{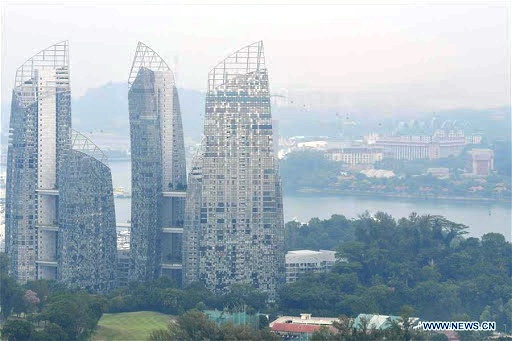}
        \label{E_3}
    }
    \\    \subfigure
    {
        \includegraphics[width=1.1in]{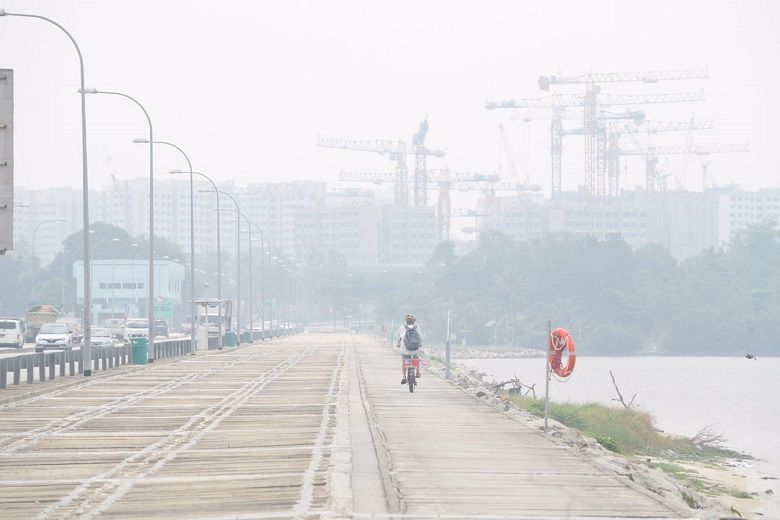}
        \label{F_1}
    }
    \subfigure
    {
        \includegraphics[width=1.1in]{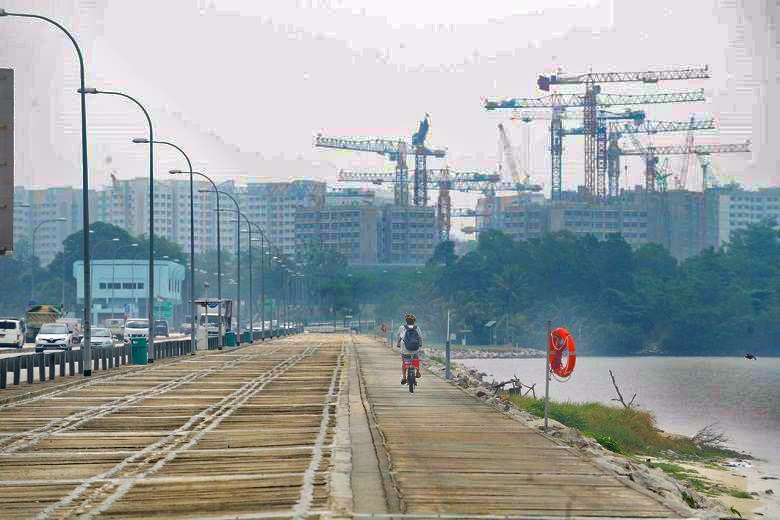}
        \label{F_2}
    }
    \subfigure
    {
        \includegraphics[width=1.1in]{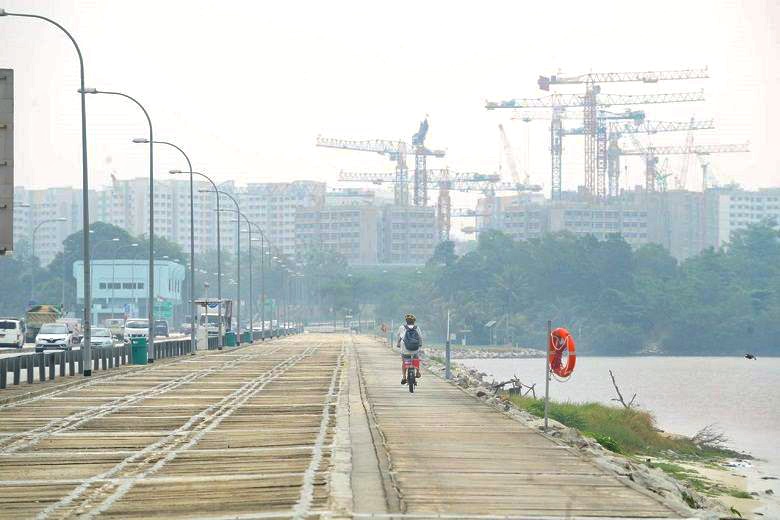}
        \label{F_3}
    }
    \caption{Visual comparison between real-world hazy images (first column), recovered images from DCP (second column) and recovered images from MLDCP (third column). }
    \label{fig:real_image}
\end{figure}

Even though synthetic benchmark datasets and metrics have achieved great success in comparing dehazing performances, the visual effect of synthetic haze is still different from the real-world haze. In Fig. \ref{fig:real_image}, the authors also present the recovered images by conventional DPC and MLDCP from real-world nature hazy scenes (first three rows) and hazy scenes in urban cities (last three rows). It is obviously observed that the MLDCP model performs better than conventional DCP in both sky and non-sky regions. The output images restored by MLDCP prevent the color distortion in sky regions, and the brightness over a global view seems more natural and closer to realistic haze-free scenes.

\subsection{Object Detection with Pre-dehazed Test Dataset}
\subsubsection{Experiment Setup}
Some object detection tasks on RESIDE RTTS dataset have been tested in \cite{li2017all} \cite{li2018benchmarking} \cite{liu2018improved}. The Domain-Adaptive Mask R-CNN model proposed in \cite{liu2018improved} achieves the highest detection accuracy on RTTS test dataset among several well-known object detection models including Faster R-CNN \cite{ren2015faster}, Mask R-CNN \cite{he2017mask}, SSD \cite{liu2016ssd} and RetinaNet \cite{lin2017focal}. \cite{li2017all} proposed a concatenation of dehazing algorithm and object detection modules to detect objects in the hazy environment. \cite{liu2018improved} further tried more combinations of effective dehazing algorithms and object detectors in the cascade and evaluated their performances on the mean average precision (mAP) values. MLDCP model is utilized to dehaze the RTTS test dataset, and Mask R-CNN and DMask-RCNN modules are fed with the recovered images. Then the authors compare the detection accuracy with the results of dehazing-detection cascades in \cite{liu2018improved}.

In Table \ref{tb:MLDCP mAP value comparasion}, the proposed cascade of MLDCP dehazing model and Mask R-CNN or Mask R-CNN increases the detection accuracy by about 2\% or 1.7\% respectively, which is the highest in comparison with other dehazing approaches, including AOD-Net \cite{li2017all}, MSCNN \cite{ren2016single} and conventional DCP \cite{he2010single}. 
And pre-dehazing RTTS dataset with MSCNN and DCP can both enhance the performance of object detection in the hazy environment. In \cite{li2017all}, AOD-Net outperforms several dehazing algorithms on PSNR and SSIM values. However, it decreases the mAP result of DMask R-CNN from 61.72\% to 60.47\%, which means that a better dehazing performance on PSNR and SSIM values does not align with a higher accuracy on object detection in the hazy environment.

\begin{table}[H]
    \centering
    \caption{\footnotesize{Object detection accuracy (mAP) comparison among different dehazing-detection cascades}}
    \begin{tabular}{c|c}
    \hline
    \textbf{Framework}  & \textbf{mAP(\%)} \\
    \hline
    \textbf{Mask R-CNN}  &  {61.01} \\
    \hline
    \textbf{DMask R-CNN2}  &  61.72 \\
    \hline
    \textbf{MLDCP + Mask R-CNN}  &  \textbf{63.06} \\
    \hline
    \textbf{AOD-Net + DMask R-CNN2}  &  60.47 \\
    \hline
    \textbf{MSCNN + DMask R-CNN2}  &  63.36 \\
    \hline
    \textbf{DCP + DMask R-CNN2}  &  \textbf{62.78} \\
    \hline
    \textbf{MLDCP + DMask R-CNN2}  & \textbf{63.42} \\
    \hline
    \end{tabular}
    
    \label{tb:MLDCP mAP value comparasion}
\end{table}

\subsection{Object Detection with Hazy-COCO Dataset}
\subsubsection{Experiment Setup}
The authors train Mask R-CNN on the new Hazy-COCO training dataset and evaluate its performance in comparison with the original Mask R-CNN trained on the MS COCO training dataset. During the training process, the backbone network is set as resnet101, and the training starts with the pre-trained COCO weights. The authors also tried various combinations of training stages to check the effect on its detection performance. It turns out that when training the head layers by ten epochs with a learning rate of 0.001 and fine-tuned layers from Resnet stage 4 and up by 60 epochs with a learning rate of 0.0001, the authors obtained the weights and bias with the highest detection accuracy on both RTTS and UG2 test datasets. 

\begin{figure}[H]
    \subfigure[]
    {
        \includegraphics[width=1.7in]{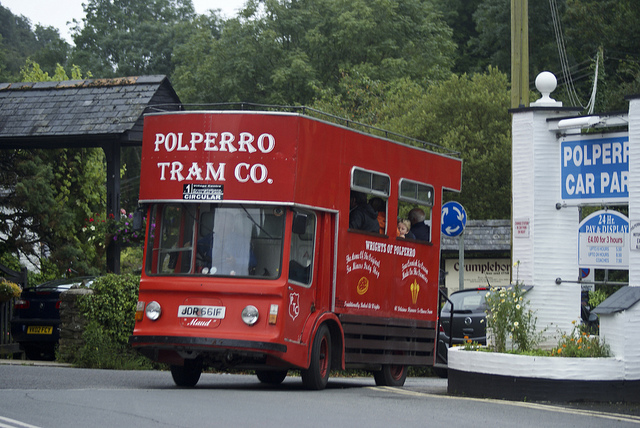}
        \label{fig4_11}
    }
    \subfigure[]
    {
        \includegraphics[width=1.7in]{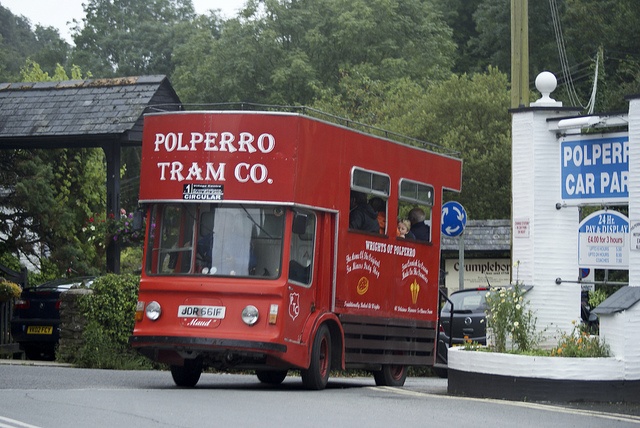}
        \label{fig4_12}
    }
    \\
    \subfigure[]
    {
        \includegraphics[width=1.7in]{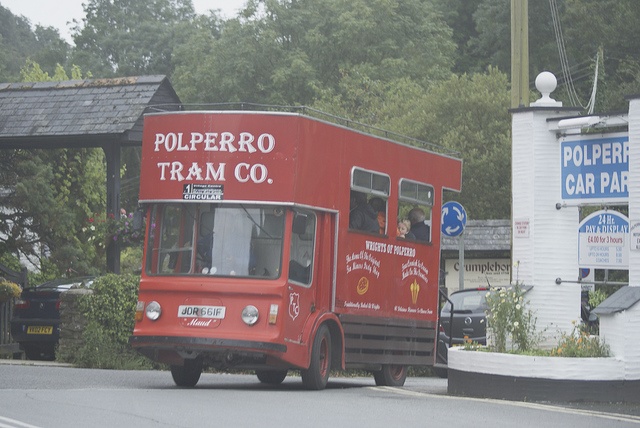}
        \label{fig4_21}
    }
    \subfigure[]
    {
        \includegraphics[width=1.7in]{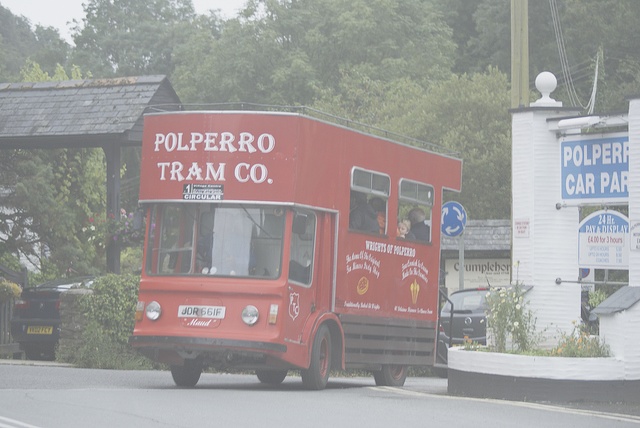}
        \label{fig4_22}
    }
    \caption{Comparison on synthetic hazy image. (a) is a haze-free image in MS COCO, (b) is the synthetic hazy image generated by inverse DCP, (c) is the synthetic hazy image with 0.1 haze density generated by our algorithm, (d) is the synthetic hazy image with 0.2 haze density generated by our algorithm.}
    \label{fig:Comparison on synthetic hazy images}
\end{figure}

\subsubsection{Experiment Results with Hazy-COCO Training Dataset}
\begin{table}[H]
    \centering
    \caption{\footnotesize{Object detection accuracy (mAP) comparison between MS COCO and Hazy-COCO training datasets}}
    \begin{tabular}{c|c|c}
    \hline
    \textbf{Test Dataset}  & \textbf{Training Dataset} & \textbf{mAP(\%)} \\
    \hline
    \textbf{RTTS}  &  \textbf{COCO} & {61.01} \\
    \hline
    {}  & \textbf{Hazy COCO} & \textbf{66.08} \\
    \hline
    \textbf{RTTS+MLDCP} & \textbf{COCO} &  {63.06} \\
    \hline
    {} & \textbf{Hazy COCO}  &  \textbf{66.15} \\
    \hline
    \textbf{UG2} & \textbf{COCO} &  {32.07} \\
    \hline
    {} & \textbf{Hazy COCO}  &  \textbf{38.53} \\
    \hline
    \end{tabular}
    
    \label{tb: Test Dataset mAP value comparasion}
\end{table}

In Table.\ref{tb: Test Dataset mAP value comparasion}, the authors evaluate the difference in detection performances when Mask R-CNN is pre-trained on MS COCO and Hazy-COCO datasets. The mAP results increase significantly by about 5\% and 6\% on RTTS and UG2 test datasets respectively, which presents how effective the Hazy-COCO training dataset is. Additionally, when training Mask R-CNN on the Hazy-COCO dataset, dehazing the RTTS test dataset by the MLDCP model can simultaneously increase the mAP result by 0.07\%. However, this increment is much lower than the increment 2\% when we preprocess the RTTS dataset individually. This is caused by the overlaps of enhancement effects between MLDCP and the advanced inverse DCP since they share some similarities over the dark channel regular pattern and multiple linear regression model.

\subsubsection{Visual Effect of Synthetic Haze}
Generally, traditional DCP does not support generating synthetic haze to a haze-free image by simply implementing the dark channel regular pattern inversely. This limitation becomes possible after the authors apply the multiple linear regression techniques to the atmospheric scattering model. 
\begin{figure}[H]
    \subfigure
    {
        \includegraphics[width=1.75in]{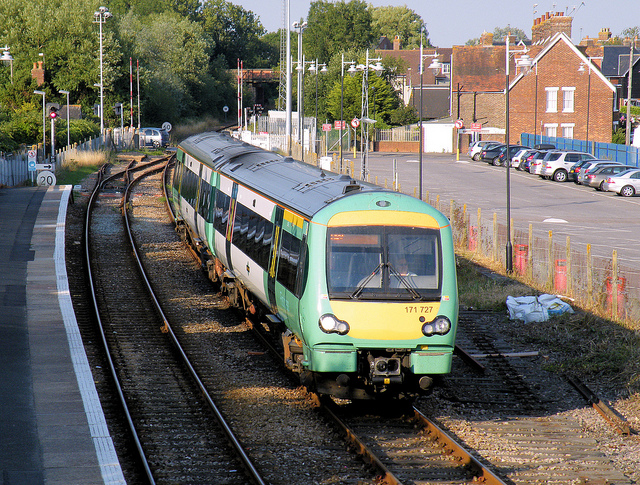}
    }
    \subfigure
    {
        \includegraphics[width=1.75in]{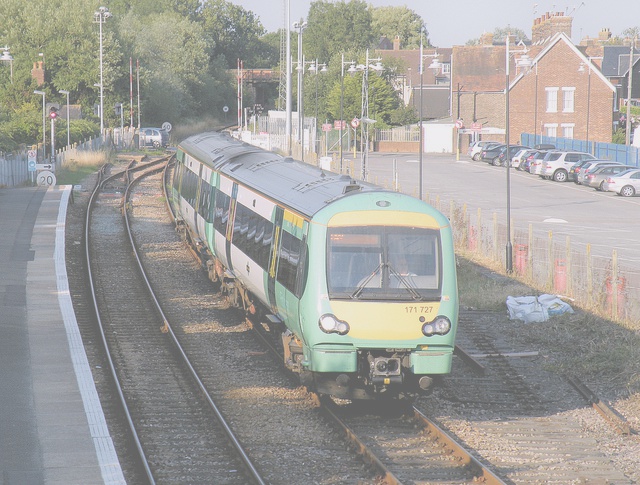}
    }
    \\
    \subfigure
    {
        \includegraphics[width=1.75in]{Figures/Fig5Clear2.jpg}
    }
    \subfigure
    {
        \includegraphics[width=1.75in]{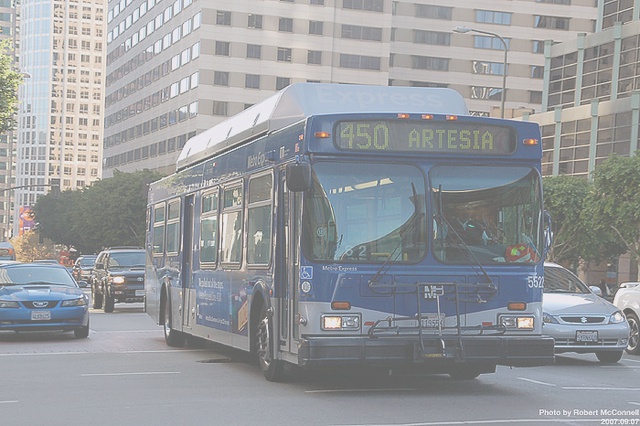}
    }
    \\
    \subfigure
    {
        \includegraphics[width=1.75in]{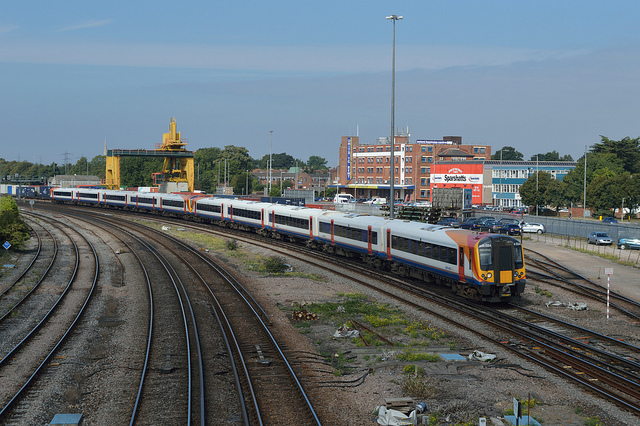}
    }
    \subfigure
    {
        \includegraphics[width=1.75in]{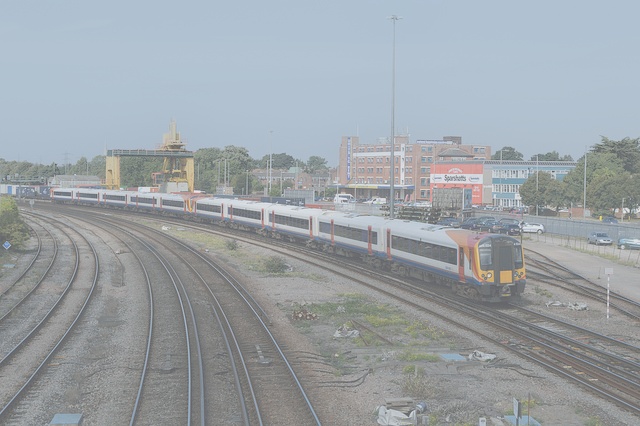}
    }
    \\
    \subfigure
    {
        \includegraphics[width=1.75in]{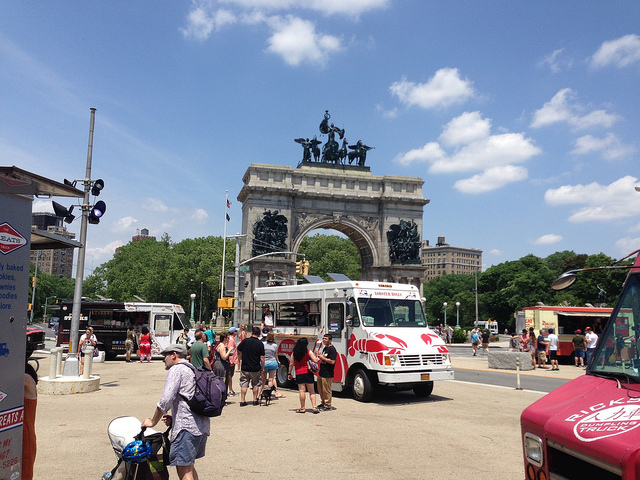}
    }
    \subfigure
    {
        \includegraphics[width=1.75in]{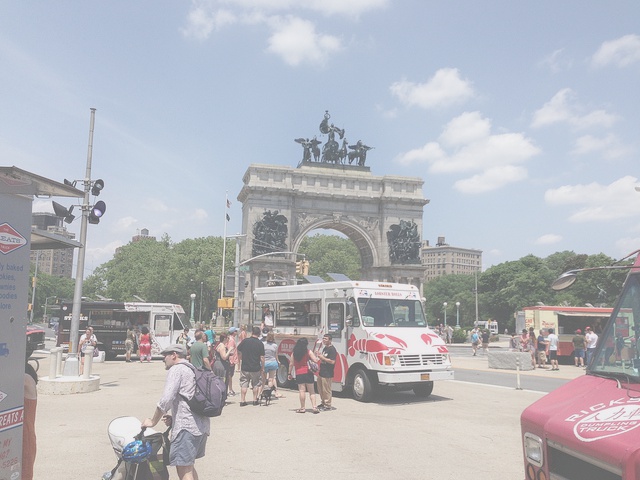}
    }
    
    \caption{Visual comparison between haze-free outdoor images from MS COCO dataset (left column) and the synthetic hazy images with 0.2 haze density generated by our algorithm (right column).}
    \label{fig:A group of synthetic hazy images}
\end{figure}

In Fig. \ref{fig:Comparison on synthetic hazy images}, the effect of haze generated by the inverse DCP algorithm can be barely observed in (b), in comparison with the haze-free image (a). Synthetic hazy images created by our advanced inverse DCP algorithm with haze density 0.1 and 0.2 are presented as (c) and (d) respectively. Both of the two synthetic hazy images are covered with seemingly realistic haze. And it is obviously observed that the synthetic haze in (d) is denser than the haze in (c). The authors also present four more groups of haze-free images from the MS COCO dataset and their corresponding synthetic hazy images in Fig. \ref{fig:A group of synthetic hazy images} to demonstrate the effectiveness of our algorithm.

\section{Conclusion}
As autonomous driving and traffic surveillance become widespread in smart cities, object detection in extreme environments like hazy has been paid special attention to. This paper proposed the approaches that increase the detection accuracy in the hazy environment from both dehazing test dataset and preprocessing training dataset.

First, an advanced Multiple Linear Regression Haze-removal Model was proposed, aiming to overcome the deficiencies of Dark Channel Prior. The authors implemented Stochastic Gradient Descent to update and find the optimal weights and bias to refine the rough estimation of two essential parameters transmission matrix $t(x)$ and atmospheric light $A$. The experimental results showed that MLDCP not only achieved higher PSNR and SSIM values than other state-of-the-art dehazing algorithms and CNN-based dehazing models, but also increased the detection precision by a higher rate when concatenated with object detection models. It demonstrated that sometimes exploiting practical conventional dehazing algorithms by machine learning techniques was superior to building more complicated neural networks. 

Second, the authors proposed an inverse DCP algorithm formulated on the multiple linear regression model that could generate synthetic haze to any existing image datasets. The authors expected this technique to prevent from spending excessive time and workload on building a large-scale image dataset with synthetic or real-world haze, as well as detailed annotations for object detection and segmentation. Synthetic haze was added to the MS COCO dataset and trained Mask R-CNN on this Hazy-COCO training set. The experimental results presented a significant increase in detection accuracy in the hazy environment. However, this approach could only generate the average synthetic haze with the same density over all pixels in an image. On the contrary, real-world haze density normally varies from pixel to pixel. In future work, it is expected to optimize this limitation and generate more realistic synthetic haze. 

\footnotesize{


\bibliographystyle{refastesj}
\bibliography{astesj.bib}

} 

\end{multicols}

\end{document}